\documentclass[11pt,letterpaper]{article}
\usepackage{ijcnlp2017}
\usepackage{times}
\usepackage{latexsym}

\ijcnlpfinalcopy

\usepackage{url}

\usepackage{multirow}
\usepackage{graphicx}
\usepackage{amssymb}
\usepackage{subfig}
\usepackage{hyperref}
\usepackage[font=footnotesize,labelfont=bf]{caption}
\usepackage{amsmath}

\usepackage{enumitem}
\setenumerate[1]{itemsep=0pt,partopsep=0pt,parsep=\parskip,topsep=2.5pt}
\setitemize[1]{itemsep=0pt,partopsep=0pt,parsep=\parskip,topsep=2.5pt}
\setdescription{itemsep=0pt,partopsep=0pt,parsep=\parskip,topsep=2.5pt}

\def\ijcnlppaperid{***}

\title{Convolutional Neural Network with Word Embeddings for\\ Chinese Word Segmentation}

\author{Chunqi Wang$^{1,2}$ \and Bo Xu$^1$ \\
	$^1$Institute of Automation, Chinese Academy of Sciences	\\
	$^2$University of Chinese Academy of Sciences	\\
	{\tt chqiwang@126.com, xubo@ia.ac.cn}}

\date{}

\begin{document}
	
	\maketitle
	
	\begin{abstract}
		Character-based sequence labeling framework is flexible and efficient for Chinese word segmentation (CWS).
		Recently, many character-based neural models have been applied to CWS. While they obtain good performance, they have two obvious weaknesses. The first is that they heavily rely on manually designed bigram feature, i.e. they are not good at capturing \emph{n}-gram features automatically. The second is that they make no use of full word information. For the first weakness, we propose a convolutional neural model, which is able to capture rich $n$-gram features without any feature engineering.
		For the second one, we propose an effective approach to integrate the proposed model with word embeddings.
		We evaluate the model on two benchmark datasets: PKU and MSR. Without any feature engineering, the model obtains competitive performance --- 95.7\% on PKU and 97.3\% on MSR. Armed with word embeddings, the model achieves state-of-the-art performance on both datasets --- 96.5\% on PKU and 98.0\% on MSR, without using any external labeled resource.
	\end{abstract}

	\section{Introduction}
	\label{sec:intro}
	
\begin{figure}
		\centering
		\includegraphics[scale=0.6]{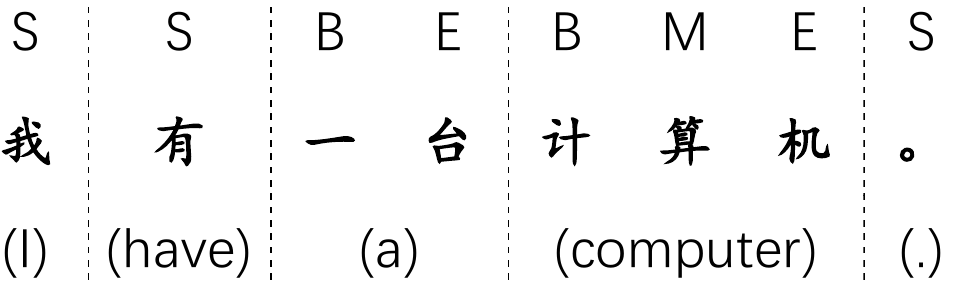}
		\caption{\label{fig:tag_scheme} Chinese word segmentation as a sequence labeling task. This figure presents the common \emph{BMES} (Begining, Middle, End, Singleton) tagging scheme.}
	\end{figure}

	Unlike English and other western languages, most east Asian languages, including Chinese, are written without explicit word delimiters. However, most natural language processing (NLP) applications are word-based. Therefore, word segmentation is an essential step for processing those languages. CWS is often treated as a character-based sequence labeling task \cite{xue2003chinese, peng2004chinese}. Figure~\ref{fig:tag_scheme} gives an intuitive explaination.
	Linear models, such as Maximum Entropy (ME) \cite{Berger1996A} and Conditional Random Fields (CRF) \cite{Lafferty2001Conditional}, have been widely used for sequence labeling tasks. However, they often depend heavily on well-designed hand-crafted features.
	
	Recently, neural networks have been widely used for NLP tasks. \newcite{Collobert2011Natural} proposed a unified neural architecture for various sequence labeling tasks. Instead of exploiting hand-crafted input features carefully optimized for each task, their system learns internal representations automatically. 
	As for CWS, there are a series of works, which share the main idea with \newcite{Collobert2011Natural} but vary in the network architecture. In particular, feed-forward neural network \cite{Zheng2013Deep}, tensor neural network \cite{Pei2014Max}, recursive neural network \cite{chen2015gated}, long-short term memory (LSTM) \cite{Chen2015Long}, as well as the combination of LSTM and recursive neural network \cite{xu2016dependency} have been used to derive contextual representations from input character sequences, which are then fed to a prediction layer.
	
	Despite of the great success of above models, they have two weaknesses. The first is that they are not good at capturing \emph{n}-gram features automatically. Experimental results show that their models perform badly when no bigram feature is explicitly used. One of the strengths of neural networks is the ability to learn features automatically. However, this strength has not been well exploited in their works.
	The second is that they make no use of full word information. Full word information has shown its effectiveness in word-based CWS systems \cite{Andrew2006A,Zhang2007Chinese,Sun2009A}. Recently, \newcite{Liu2016Exploring,Zhang2016Transition} utilized word embeddings to boost performance of word-based CWS models. However, for character-based CWS models, word information is not easy to be integrated. 
	
	For the first weakness, we propose a convolutional neural model, which is also character-based.  Previous works have shown that convolutional layers have the ablity to capture rich \emph{n}-gram features \cite{kim2015character-aware}. We use stacked convolutional layers to derive contextual representations from input sequence, which are then fed into a CRF layer for sequence-level prediction.
	For the second weakness, we propose an effective approach to incorporate word embeddings into the proposed model. The word embeddings are learned from large auto-segmented data. Hence, this approach belongs to the category of semi-supervised learning.
	
	We evaluate our model on two benchmark datasets: PKU and MSR.
	Experimental results show that even without the help of explicit $n$-gram feature, our model is capable of capturing rich \emph{n}-gram information automatically, and obtains competitive performance --- 95.7\% on PKU and 97.3\% on MSR (F score). Furthermore, armed with word embeddings, our model achieves state-of-the-art performance on both datasets --- 96.5\% on PKU and 98.0\% on MSR, without using any external labeled resource. \footnote{The tensorflow \cite{abadi2016tensorflow} implementation and related resources can be found at \url{https://github.com/chqiwang/convseg}.}
	
	\section{Architecture}
	
	In this section, we introduce the architecture from bottom to top.
	\subsection{Lookup Table}
	\label{sec:arc/lookup}
	The first step to process a sentence by deep neural networks is often to transform words or characters into embeddings \cite{Bengio2003A, Collobert2011Natural}. This transformation is done by lookup table operation.
	A character lookup table $M_{char} \in \mathbb{R}^{|V_{char}| \times d}$ (where $|V_{char}|$ denotes the size of the character vocabulary and $d$ denotes the dimension of embeddings) is associated with all characters.
	Given a sentence $\mathcal{S} = (c_1, c_2, ..., c_L)$, after the lookup table operation, we obtain a matrix $X \in \mathbb{R}^{L \times d}$ where the $i$'th row is the character embedding of $c_i$.
	
	Besides the character, other features can be easily incorporated into the model (we shall see word feature in section \ref{sec:wemb}). We associate to each feature a lookup table (some features may share the same lookup table) and the final representation is calculated as the concatenation of all corresponding feature embeddings. 
	
	\subsection{Convolutional Layer}
	
	\begin{figure}
		\centering
		\includegraphics[scale=0.35]{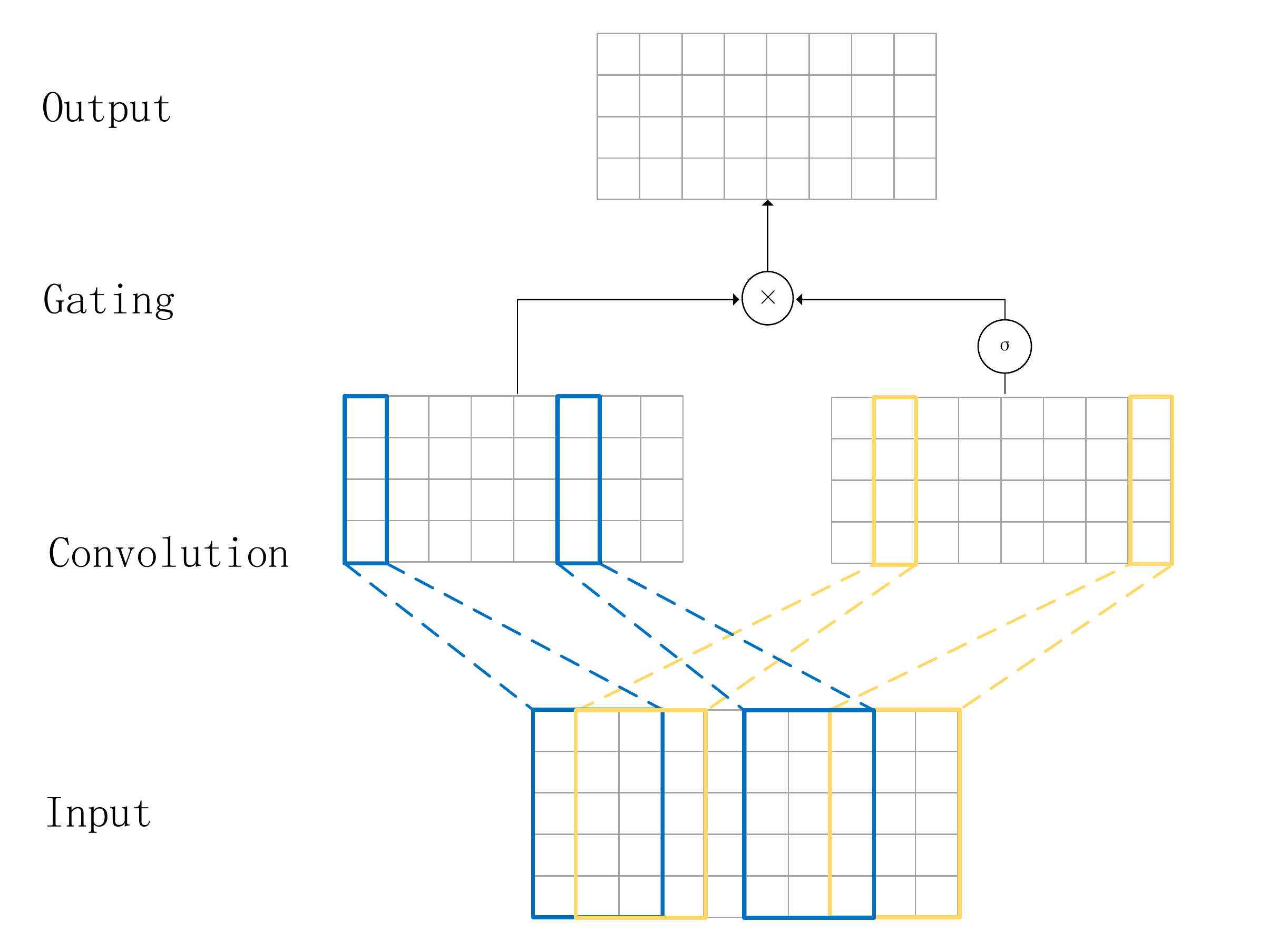}
		\caption{Structure of a convolutional layer with GLU. There are five input channels and four output channels in this figure.}
		\label{fig:convlayer}
	\end{figure}
	
	Many neural network models have been explored for CWS. However, experimental results show that they are not able to capure $n$-gram information automatically \cite{Pei2014Max,chen2015gated,Chen2015Long}. To achieve good performance, $n$-gram feature must be used explicitly. 
	To overcome this weakness, we use convolutional layers \cite{waibel1989phoneme} to encode contextual information.
	Convolutional neural networks (CNNs) have shown its great effectiveness in computer vision tasks \cite{krizhevsky2012imagenet,simonyan2014very,He_2016_CVPR}. Recently,
	\newcite{zhang2015character} applied character-level CNNs to text classification task. They showed that CNNs tend to outpeform traditional $n$-gram models as the dataset goes larger. \newcite{kim2015character-aware} also observed that character-level CNN learns to differentiate between different types of $n$-grams --- prefixes, suffixes and others, automatically.
	
	Our network is quite simple --- only convolutional layers is used (no pooling layer).
	Gated linear unit (GLU) \cite{Dauphin2016Language} is used as the non-linear unit in our convolutional layer, which has been shown to surpass rectified linear unit (ReLU) on the language modeling task. For simplicity, GLU can also be easily replaced by ReLU with performance slightly hurt (with roughly the same number of network parameters). Figure~\ref{fig:convlayer} shows the structure of a convolutional layer with GLU.
	Formally, we define the number of input channels as $N$, the number of output channels as $M$, the length of input as $L$ and kernel width as $k$. The convolutional layer can be written as
	$$
	F(X)=(X*W+b) \otimes \sigma(X*V+c)
	$$
	where $*$ denotes one dimensional convolution operation,  $X \in \mathbb{R}^{L\times N}$ is the input of this layer, $W \in \mathbb{R}^{k\times N\times M}$, $b \in \mathbb{R} ^{M}$, $V \in \mathbb{R}^{k\times N\times M}$, $c \in \mathbb{R} ^{M}$ are parameters to be learned, $\sigma$ is the sigmoid function and $\otimes$ represents element-wise product.
	We make $F(X) \in \mathbb{R}^{L \times M}$ by augmenting the input $X$ with paddings.
	
\begin{figure}
		\centering
		\includegraphics[scale=0.4]{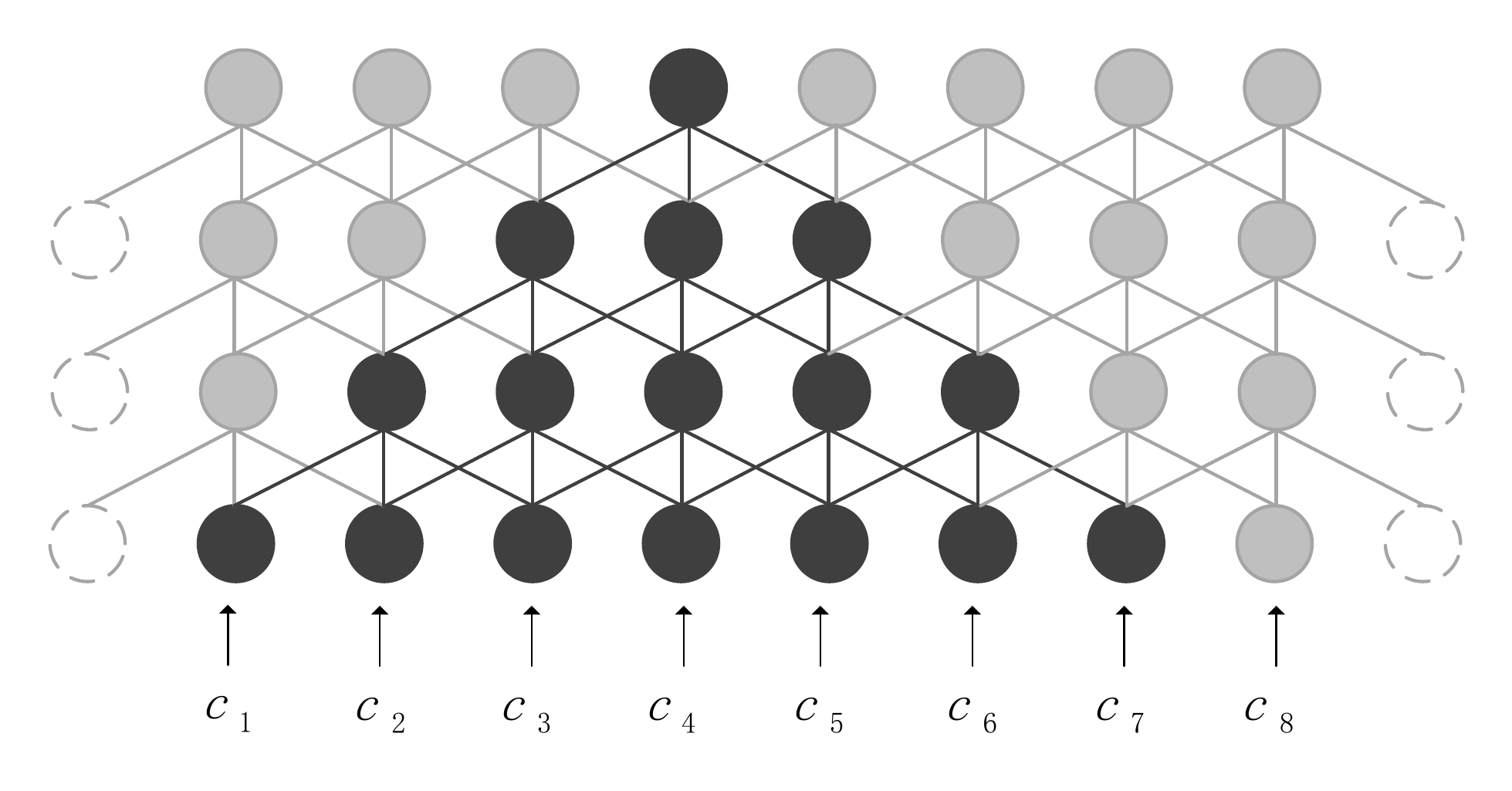}
		\caption{Stacked convolutional layers. There is one input layer on the bottom and three convolutional layers on the top. Dashed white circles denote paddings. Black circles and lines mark the pyramid in the perspective of $c_4$.}
		\label{fig:stackedconv}
	\end{figure}

	A succession of convolutional layers are stacked to capture long distance information. From the perspective of each character, information flows in a pyramid. Figure~\ref{fig:stackedconv} shows a network with three convolutional layers stacked.
	On the topmost layer, a linear transformation is used to transform the output of this layer to unnormalized label scores $E \in \mathbb{R}^{L \times C}$, where $C$ is the number of label types.
	
	\subsection{CRF Layer}
	
	For sequence labeling tasks, it is often beneficial to explicitly consider the correlations between adjacent labels \cite{Collobert2011Natural}.
	
	Correlations between adjacent labels can be modeled as a transition matrix $T \in \mathbb{R}^{C\times C}$. Given a sentence 
	$\mathcal{S}=(c_1, c_2,...,c_L)$,
	we have corresponding scores $E \in \mathbb{R}^{L \times C}$ given by the convolutional layers.
	For a label sequence $y=(y_1,y_2,...,y_L)$, we define its unnormalized score to be
	$$
	s(\mathcal{S},y)=\sum\limits_{i=1}^{L}{E_{i, y_i}} + \sum_{i=1}^{L-1}{T_{y_i,y_{i+1}}}.
	$$
	then the probability of the label sequence is defined as
	$$
	P(y|\mathcal{S})=\frac{e^{s(\mathcal{S},y)}}{\sum_{y\prime \in \mathcal{Y}}e^{s(\mathcal{S},y\prime)}}
	$$
	where $\mathcal{Y}$ is the set of all valid label sequences. 
	This actually takes the form of linear chain CRF \cite{Lafferty2001Conditional}.
	Then the final loss of the proposed model is defined as the negative log-likehood of the ground-truth label sequence $y^*$
	$$
	\mathcal{L}(\mathcal{S},y^\star)  = -log P(y^\star|\mathcal{S}).
	$$
	
	During training, the loss function is minimized by back propagation. During test, Veterbi algorithm is applied to quickly find the label sequence with maximum probability.
	
	\section{Integration with Word Embeddings}
	\label{sec:wemb}
	
	Character-based CWS models have the superiority of being flexible and efficient. However, full word information is not easy to be incorporated. 
	There is another type of CWS models: the word-based models. Models belong to this category utilize not only character-level information, but also word-level \cite{Zhang2007Chinese,Andrew2006A,Sun2009A}. 
	Recent works have shown that word embeddings learned from large auto-segmented data lead to great improvements in word-based CWS systems \cite{Liu2016Exploring,Zhang2016Transition}. 
	We propose an effective approach to integrate word embeddings with our character-based model. The integration brings two benefits. On the one hand, full word information can be used. On the other hand, large unlabeled data can be better exploited.
	
	\begin{table}
		\begin{tabular}{c|l}
			\hline
			Length	&	Features	\\
			\hline
			1	&	$\underline{c_i}$	\\
			\hline
			2	&	$c_{i-1}\underline{c_i} \quad \underline{c_i}c_{i+1}$	\\
			\hline
			3	&	$c_{i-2}c_{i-1}\underline{c_i} \quad c_{i-1}\underline{c_i}c_{i+1} \quad  \underline{c_i}c_{i+1}c_{i+2}$	\\
			\hline
			\multirow{2}{*}{4}
			&	$c_{i-3}c_{i-2}c_{i-1}\underline{c_i} \quad c_{i-2}c_{i-1}\underline{c_i}c_{i+1}$	\\
			&	$c_{i-1}\underline{c_i}c_{i+1}c_{i+2} \quad \underline{c_i}c_{i+1}c_{i+2}c_{i+3}$	\\
			\hline
		\end{tabular}
		\caption{\label{tab:wordfeature}Word features at position $i$ given a sentence $\mathcal{S}=(c_1, c_2,...,c_L)$. Only the words that include the current character $c_i$ (marked with underline) are considered as word feature. Hence, the number of features can be controlled in a reasonable range. We also restrict the max length of words to 4 since few words contain more than 4 characters in Chinese. Note that the feature space is still tremendous ($O(N^4)$, where $N$ is the number of characters).}
	\end{table}
	
	To use word embeddings, we design a set of word features, which are listed in Table~\ref{tab:wordfeature}. We associate to the word features a lookup table $M_{word}$. Then the final representation of $c_i$ is defined as
	\begin{equation*}
		\begin{split}
			R(c_i) = & M_{char}[c_i] \oplus\\ 
			& M_{word}[c_i] \oplus M_{word}[c_{i-1}c_i] \oplus \dots \oplus \\
			& M_{word}[c_ic_{i+1}c_{i+2}c_{i+3}]
		\end{split}
	\end{equation*}
	where $\oplus$ denotes the concatenation operation. 
	Note that the max length of word features is set to 4, therefore the feature space is extremely large ($O(N^4)$). 
	A key step is to shrink the feature space so that the memory cost can be confined within a feasible scope. In the meanwhile, the problem of data sparsity can be eased. The solution is as following.
	Given the unlabeled data $\mathcal{D}_{un}$ and a teacher CWS model, we segment $\mathcal{D}_{un}$ with the teacher model and get the auto-segmented data $\mathcal{D}_{un}^\prime$. A vocabulary $V_{word}$ is generated from $\mathcal{D}_{un}^\prime$ where low frequency words are discarded \footnote{The threshold of frequency is set to 5, which is the default setting of \emph{word2vec}.}.
	We replace $M_{word}[*]$ with $M_{word}[\emph{UNK}]$ if $* \notin V_{word}$ ($\emph{UNK}$ denotes the unknown words).
	
	To better exploit the auto-segmented data $\mathcal{D}_{un}^\prime$, we adopt an off-the-self tool \emph{word2vec}\footnote{\url{https://code.google.com/p/word2vec}} \cite{mikolov2013efficient} to pretrain the word embeddings.
	The whole procedure is summarized as following setps:
	\begin{enumerate}
		\item Train a \emph{teacher} model that does not rely on word feature.
		\item Segment unlabeled data $\mathcal{D}$ with the \emph{teacher} model and get the auto-segmented data $\mathcal{D}^\prime$.
		\item Build a vocabulary $V_{word}$ from $\mathcal{D}^\prime$. Replace all words not appear in $V_{word}$ with \emph{UNK}.
		\item Pretrain word embeddings on $\mathcal{D}^\prime$ using \emph{word2vec}.
		\item Train the \emph{student} model with word feature using the pretrained word embeddings.
	\end{enumerate}
	
	Note that no external labeled data is used in this procedure.
	
	\section{Experiments}
	
	\subsection{Settings}
	\noindent\textbf{Datasets}\quad 
	We evaluate our model on two benchmark datasets, PKU and MSR, from the second International Chinese Word Segmentation Bakeoff \cite{emerson2005second}. Both datasets are commonly used by previous state-of-the-art models. For both datasets, last 10\% of the training data are used as development set.
	The unlabeled data used in this work is news data collected by \emph{Sogou} \footnote{\url{http://www.sogou.com/labs/resource/ca.php}}.
	
	We do not perform any preprocessing for these datasets, such as replacing continuous digits and English characters with a single token.
	
	\noindent\textbf{Dropout}\quad
	Dropout \cite{srivastava2014dropout} is a very efficient method for preventing overfit, especially when the dataset is small. We apply dropout to our model on all convolutional layers and embedding layers. The dropout rate is fixed to 0.2.
	
	\noindent\textbf{Hyper-parameters}\quad
	For both datasets, we use the same set of hyper-parameters, which are presented in Table~\ref{tab:hyper_params}. For all convolutional layers, we just use the same number of channels. Following the practice of designing very deep CNN in computer vision \cite{simonyan2014very}, we use a small kernal width, i.e. 3,  for all convolutional layers. To avoid computational inefficiency, we use a relatively small dimension, i.e. 50, for word embeddings.
	
	\begin{table}
		\centering
		\begin{tabular}{l|c}
			\hline
			Hyper-parameters	&	Value	\\
			\hline
			dimension of character embedding	&	200	\\
			dimension of word embedding	&	50	\\
			number of conv layers	&	5	\\
			number of channels per conv layer	&	200	\\
			kernel width of filters	&	3	\\
			dropout rate	&	0.2	\\
			\hline
		\end{tabular}
		\caption{\label{tab:hyper_params} Hyper-parameters we choose for our model.}
	\end{table}
	
	\begin{table*}
		\centering
		\begin{tabular}{l|ccc|ccc}
			\hline
			\multirow{2}{*}{Models}	&	\multicolumn{3}{c|}{PKU}	&	\multicolumn{3}{c}{MSR}	\\
			\cline{2-7}
			&	P	&	R	&	F	&	P	&	R	&	F	\\
			\hline
			\cite{tseng2005conditional}	&	94.6	&	95.4	&	95.0	&	96.2	&	96.6	&	96.4	\\
			\cite{Zhang2007Chinese}	&	-	&	-	&	94.5	&	-	&	-	&	97.2	\\
			\cite{zhao2011integrating}	&	-	&	-	&	95.40	&	-	&	-	&	97.58	\\
			\cite{Sun2012Fast}	&	95.8	&	94.9	&	95.4	&	97.6	&	97.2	&	97.4	\\
			\cite{zhang2013exploring}	&	96.5	&	95.8	&	96.1	&	-	&	-	&	97.45	\\
			\cite{Pei2014Max}	&	-	&	-	&	95.2	&	-	&	-	&	97.2	\\
			\cite{chen2015gated}$^{*\diamond}$	&	96.5	&	\bf	96.3	&	96.4	&	97.4	&	97.8	&	97.6	\\
			\cite{Chen2015Long}$^{*\diamond}$	&	96.6	&	\bf	96.4	&	\bf	96.5	&	97.5	&	97.3	&	97.4	\\
			\cite{Cai2016Neural}$^\diamond$	&	95.8	&	95.2	&	95.5	&	96.3	&	96.8	&	96.5	\\
			\cite{Liu2016Exploring}	&	-	&	-	&	95.67	&	-	&	-	&	97.58	\\
			\cite{Zhang2016Transition}	&	-	&	-	&	95.7	&	-	&	-	&	97.7	\\
			\cite{xu2016dependency}$^{*\diamond}$		&	-	&	-	&	96.1	&	-	&	-	&	96.3	\\
			\hline
			\hline
			CONV-SEG	&	96.1	&	95.2	&	95.7	&	97.4	&	97.3	&	97.3	\\
			WE-CONV-SEG (+ word embeddings)	&	\bf	96.8	&	96.1	&	\bf	96.5	&	\bf 97.9	&	\bf 98.1	&	\bf 98.0	\\
			\hline
		\end{tabular}
		\caption{\label{tab:results} Performance of our models and previous state-of-the-art models. Note that \cite{chen2015gated,Chen2015Long,xu2016dependency} used a external Chinese idiom dictionary. To make the comparison fair, we mark them with $^*$. \newcite{chen2015gated,Chen2015Long,Cai2016Neural,xu2016dependency} also preprocessed the datasets by replacing the conitinous English character and digits with a unique token. We mark them with $^\diamond$.}
	\end{table*}
	
	\noindent\textbf{Pretraining}\quad
	Character embeddings and word embeddings are pretrained on unlabeled or auto-segmented data by \emph{word2vec}. Since the pretrained embeddings are not task-oriented, they are fine-tuned during supervised training by normal backpropagation.\footnote{We also try to use fixed word embeddings as \newcite{Zhang2016Transition} do but no significant difference is observed.}
	
	\noindent\textbf{Optimization}\quad
	Adam algorithm \cite{kingma2014adam} is applied to optimize our model. We use default parameters given in the original paper and we set batch size to 100.  For both datasets, we train no more than 100 epoches. The final models are chosen by their performance on the development set.
	
	Weight normalization \cite{Salimans2016Weight} is applied for all convolutional layers to accelerate the training procedure and obvious acceleration is observed.
	
	\subsection{Main Results}
	Table~\ref{tab:results} gives the performances of our models, as well as previous state-of-the-art models. 
	Two proposed models are shown in the table:
	\begin{itemize}
		\item {\bf CONV-SEG} \quad It is our preliminary model without word embeddings. Character embeddings are pretrained on large unlabeled data.
		\item {\bf WE-CONV-SEG} \quad On the basis of CONV-SEG, word embeddings are used. We use CONV-SEG as the teacher model (see section~\ref{sec:wemb}).
	\end{itemize}
	\begin{figure}
		\centering
		\subfloat{\includegraphics[scale=0.3]{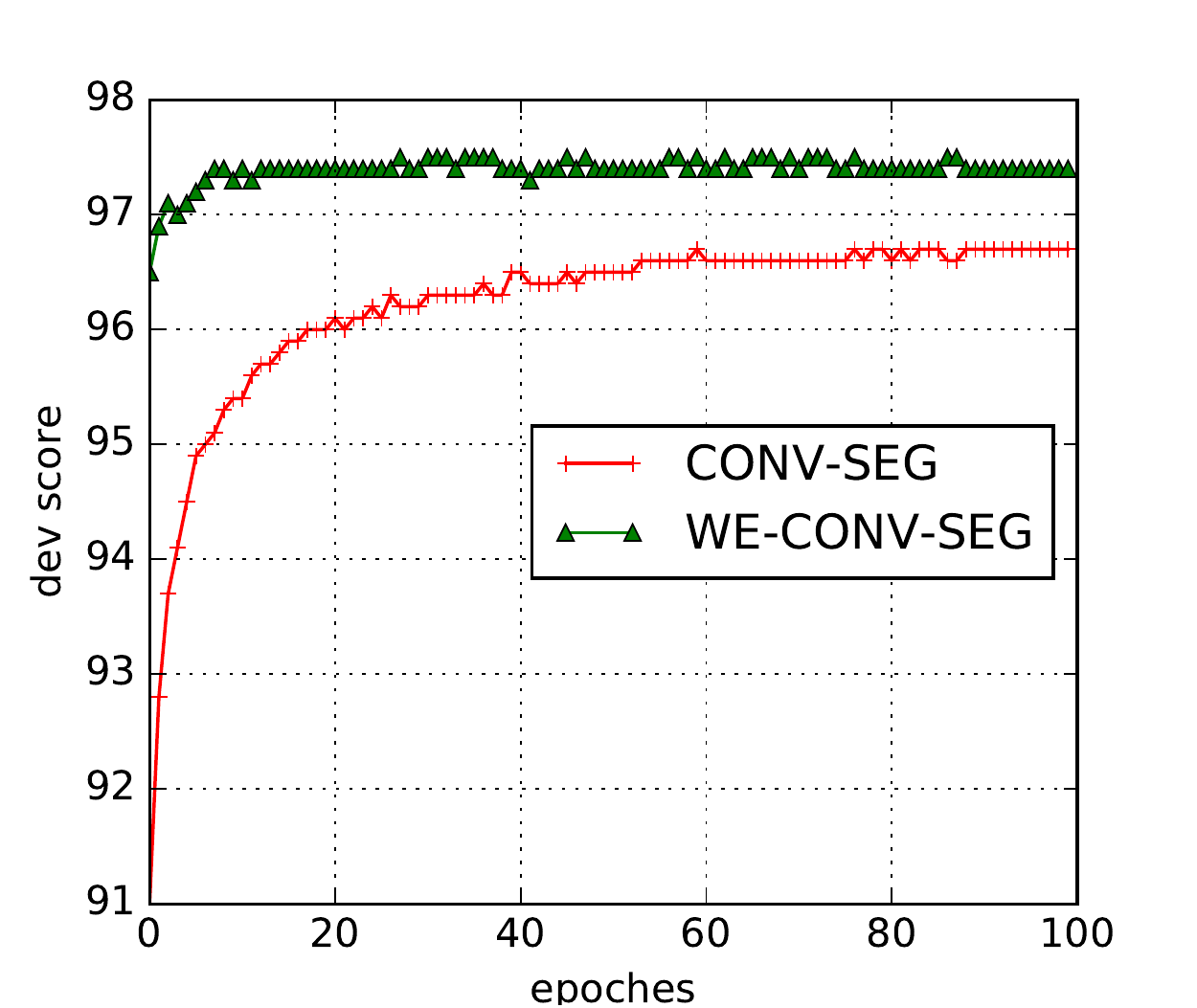}}
		\subfloat{\includegraphics[scale=0.3]{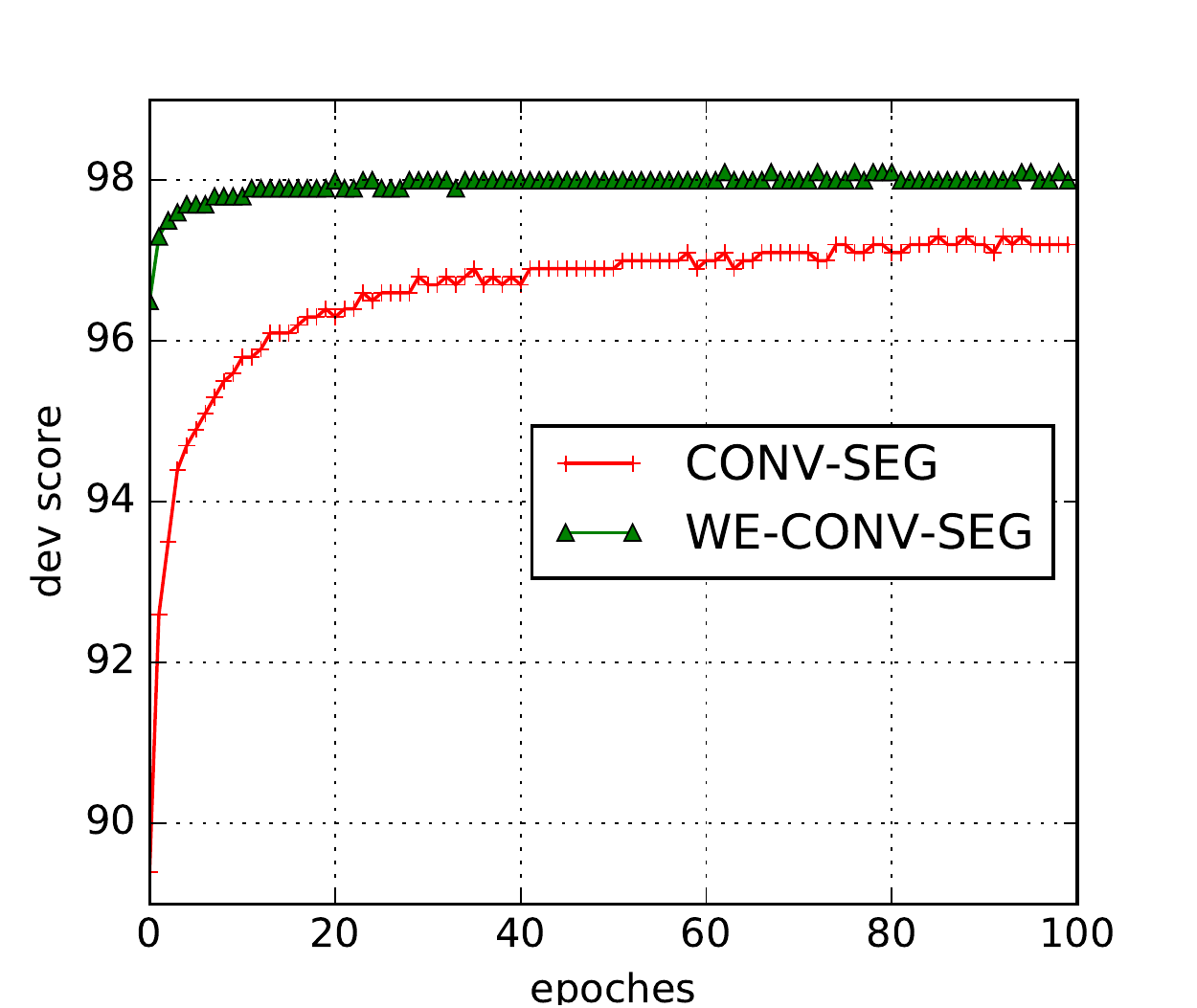}}
		\caption{\label{fig:curves}Learning curves (dev scores) of our models on PKU (left) and MSR (right).}
	\end{figure}
	
	Our preliminary model CONV-SEG achieves competitive performance without any feature engineering. Armed with word embeddings, WE-CONV-SEG obtains state-of-the-art performance on both PKU and MSR datasets without using any external labeled data. WE-CONV-SEG outperforms state-of-the-art neural model \cite{Zhang2016Transition} in a large margin (+0.8 on PKU and +0.3 in MSR). \newcite{Chen2015Long} preprocessed all datasets by replacing Chinese idioms with a single token and thus their model obtains excellent score on PKU dataset. However, WE-CONV-SEG achieves the same performance on PKU and outperforms their model on MSR, without any data preprocessing.
	
	We also observe that WE-CONV-SEG converges much faster compared to CONV-SEG. 
	Figure~\ref{fig:curves} presents the learning curves of the two models. It takes 10 to 20 epoches for WE-CONV-SEG to converge while it takes more than 60 epoches for CONV-SEG to converge.

	\subsection{Network Depth}
\begin{figure}
		\centering
		\includegraphics[scale=0.6]{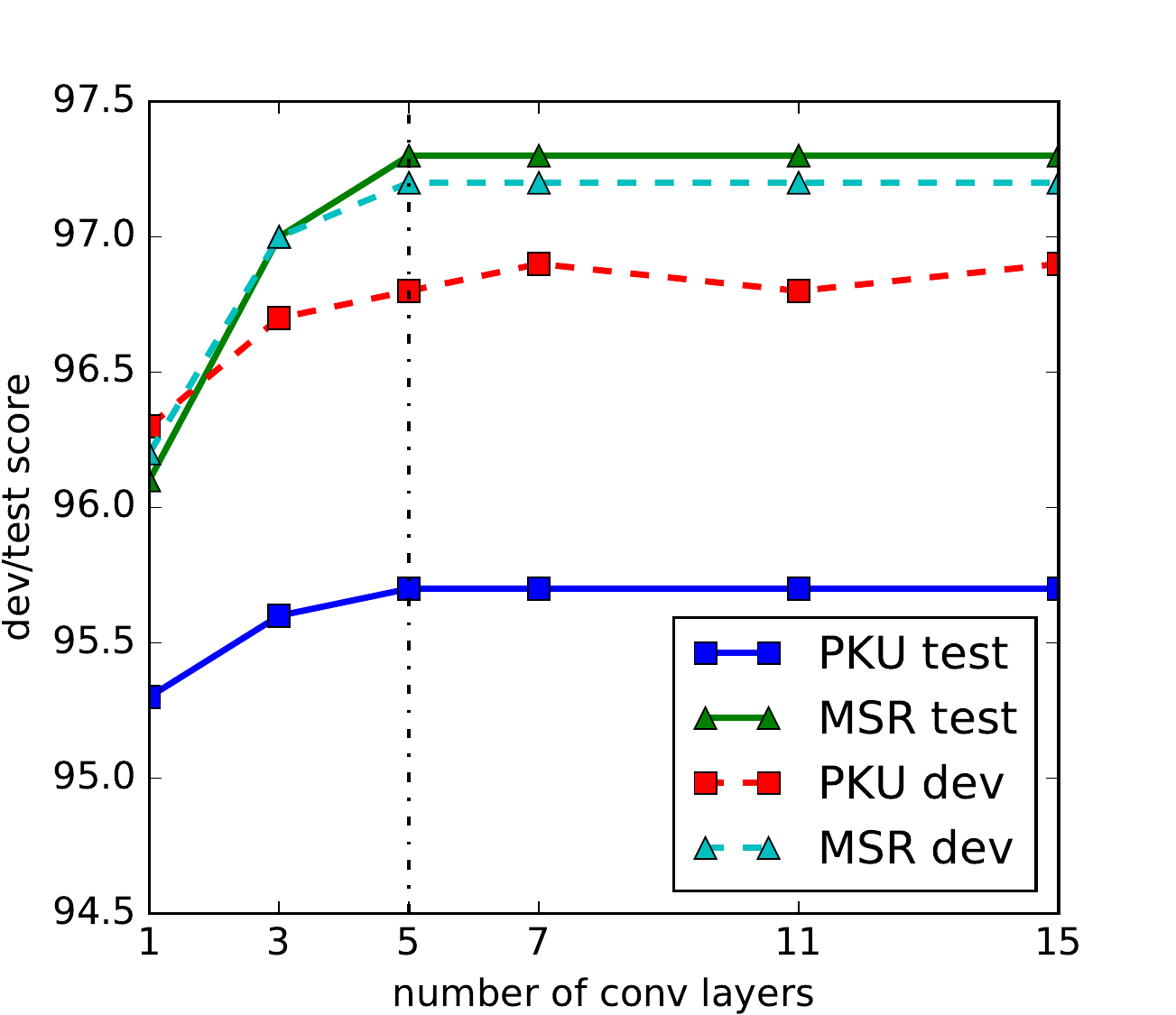}
		\caption{Scores on dev set and test set with respect to the number of convolutional layers. The vertical dashed line marks the depth we choose.}
		\label{fig:depth}
	\end{figure}

	Network depth shows great influence on the performance of deep neural networks. A too shallow network may not fit the training data very well while a too deep network may overfit or is hard to train. We evaluate the performance of the proposed model with varying depth. Figure~\ref{fig:depth} shows the results. It is obvious that five convolutional layers is a good choise for both datasets. When we increase the depth from 1 to 5, the performance is improved significantly. However, when we increase depth from 5 to 7, even to 11 and 15, the performance is almost unchanged. 
	This phenomenon implies that CWS rarely relies on context larger than 11 \footnote{Context size is calculated by $(k - 1) \times d + 1$, where $k$ and $d$ denotes the kernel size and the number of convolutional layers, respectively.}.
	With more training data, deeper networks may perform better.
	
	\subsection{Pretraining Character Embeddings}
	\begin{table}
		\centering
		\begin{tabular}{l|c|c}
			\hline
			Model Options	&	PKU	&	MSR	\\
			\hline
			without pretraining	&	94.7	&	96.7	\\
			with pretraining	&	\bf	95.7	&	\bf	97.3	\\
			\hline
		\end{tabular}
		\caption{\label{tab:pretrain}Test performances with or without pretraining character embeddings. ``without pretraining'' means that the character embeddings are randomly initialized. }
	\end{table}
	
	Previous works have shown that pretraining character embeddings boost the performance of neural CWS models significantly \cite{Pei2014Max,chen2015gated,Chen2015Long,Cai2016Neural}. We verify this and get a consistent conclusion. Table~\ref{tab:pretrain} shows the performances with or without pretraining. 
	Our model obtains significant improvements (+1.0 on PKU and +0.6 on MSR) with pretrained character embeddings.
	
	\subsection{\emph{N}-gram Features}
	
	\begin{table}
		\centering
		\resizebox{78mm}{!}{
			\begin{tabular}{l|cc|cc}
				\hline
				Models	&	PKU	&	MSR	&	PKU	&	MSR	\\
				\hline
				\cite{Cai2016Neural}$^\diamond$	&	95.5	&	96.5	&	-	&	-	\\
				\cite{Zheng2013Deep}	&	92.8$^\ddagger$	&	93.9$^\ddagger$	&	-	&	-	\\
				\cite{Pei2014Max}	&	94.0	&	94.9	&	-	&	-	\\	
				\cite{chen2015gated}	&	94.5$^\dagger$	&	95.4$^\dagger$	&	96.1$^*$	&	96.2$^*$	\\
				\cite{Chen2015Long}	&	94.8$^\dagger$	&	95.6$^\dagger$	&	96.0$^*$	&	96.6$^*$	\\
				\cite{xu2016dependency}	&	-	&	-	&	96.1$^*$	&	96.3$^*$	\\
				CONV-SEG	&	95.7	&	97.3	&	-	&	-	\\
				\hline
				\hline
				\multirow{2}{*}{\cite{Pei2014Max}}	&	95.2	&	97.2	&	\multirow{2}{*}{-}	&	\multirow{2}{*}{-}	\\	
				&	(+1.2)	&	(+2.3)	&&	\\
				\multirow{2}{*}{\cite{chen2015gated}}	&	\multirow{2}{*}{-}	&	\multirow{2}{*}{-}	&	96.4$^*$	&	97.6$^*$	\\
				&&&	(+0.3)	&	(+1.4)	\\
				\multirow{2}{*}{\cite{Chen2015Long}}	&	\multirow{2}{*}{-}	&	\multirow{2}{*}{-}	&	96.5$^*$	&	97.3$^*$	\\
				&&&	(+0.5)	&	(+0.7)	\\
				\multirow{2}{*}{AVEBE-CONV-SEG}	&	95.4	&	97.1	&	\multirow{2}{*}{-}	&	\multirow{2}{*}{-}	\\
				&	(-0.3)	&	(-0.2)	&& \\
				W2VBE-CONV-SEG	&	95.9	&	97.5	&	\multirow{2}{*}{-}	&	\multirow{2}{*}{-}	\\
				&	(+0.2)	&	(+0.2)	&&	\\
				\hline
			\end{tabular}}
			\caption{\label{tab:nobigram}The first/second group summarize results of models without/with bigram feature. The number in the parentheses is the absolute improvement given by explicit bigram feature. Results with $^*$ used external dictionary. Results with $^\dagger$ come from \newcite{Cai2016Neural}. Results with $^\ddagger$ come from \newcite{Pei2014Max}. $^\diamond$ marks word-based models.}
		\end{table}
		In this section, we test the ability of our model in capturing \emph{n}-gram features. Since unigram is indispensable and trigram is beyond memory limit, we only consider bigram.
		
		Bigram feature has shown to play a vital role in character-based neural CWS models \cite{Pei2014Max,chen2015gated,Chen2015Long}. Without bigram feature, previous models perform badly. Table~\ref{tab:nobigram} gives a summarization. Without bigram feature, our model outperforms previous character-based models in a large margin (+0.9 on PKU and +1.7 on MSR). Compared with word-based model \cite{Cai2016Neural}, the improvements are also significant (+0.2 on PKU and +0.8 on MSR).
		
		Then we arm our model with bigram feature. The bigram feature we use is the same with \newcite{Pei2014Max,chen2015gated,Chen2015Long}. The dimension of bigram embedding is set to 50.
		Following \newcite{Pei2014Max,chen2015gated,Chen2015Long}, the bigram embeddings are initialized by the average of corresponding pretrained character embeddings. The result model is named AVEBE-CONV-SEG and the performance is shown in Table~\ref{tab:nobigram}.
		Unexpectedly, the performance of AVEBE-CONV-SEG is worse than the preliminary model CONV-SEG that uses no bigram feature (-0.3 on PKU and -0.2 on MSR). This result is dramatically inconsistent with previous works, in which the performance is significantly improved by the method. 
		We also observe that the training cost of AVEBE-CONV-SEG is much lower than CONV-SEG. Hence we can conclude that the inconsistency is casued by overfitting. A reasonable conjecture is that the model CONV-SEG already capture abundant bigram feature automatically, therefore the model is tend to overfit when bigram feature is explicitly added.
		
		A practicable way to overcome overfitting is to introduce priori knowledge. We introduce priori knowledge by using bigram embeddings directly pretrained on large unlabeled data, which is simmillar with \cite{mansur2013feature}.
		We convert the unlabeled text to bigram sequence and then apply \emph{word2vec} to pretrain the bigram embeddings directly. The result model is named W2VBE-CONV-SEG, and the performance is also shown in Table~\ref{tab:nobigram}.
		This method leads to substantial improvements (+0.5 on PKU and +0.4 MSR) over AVEBE-CONV-SEG. However, compared to CONV-SEG, there are only slight gains (+0.2 on PKU and MSR). 
		
		All above observations verify that our proposed network has considerable superiority in capturing \emph{n}-gram, at least bigram features automatically.
		
		\subsection{Word Embeddings}
\begin{figure}[t]
			\centering
			\includegraphics[scale=0.45]{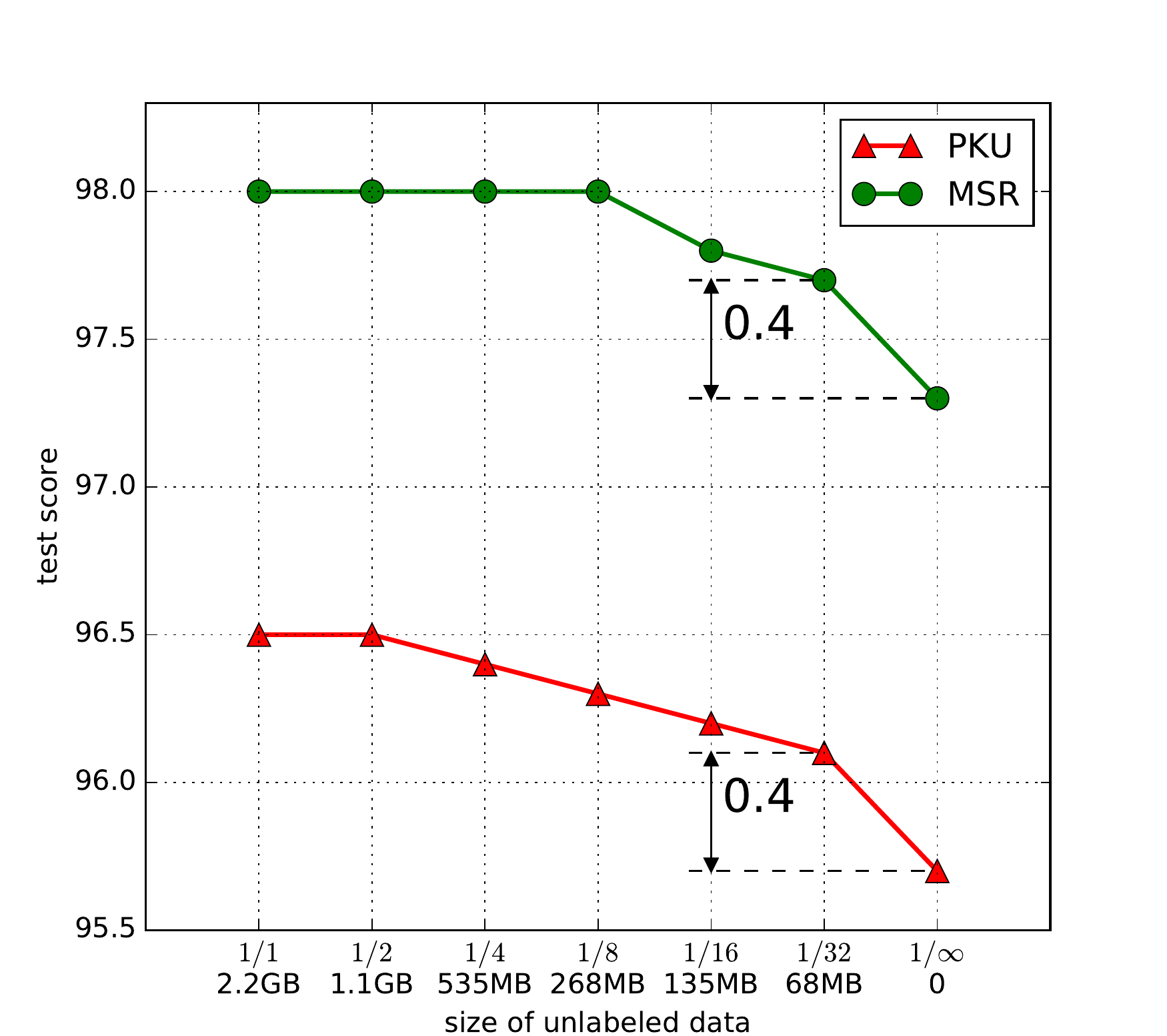}
			\caption{\label{fig:wordemb} Test performances with varying size of unlabeled data for pretraining word embeddings. With full size, the model is WE-CONV-SEG. With the 0 size, the model degenerates to CONV-SEG.}
		\end{figure}

		Word embeddings lead to significant improvements over the strong baseline model CONV-SEG. The improvements come from the teacher model and the large unlabeled data.
		A natural question is how much unlabeled data can lead to significant improvements. We study this by halving the unlabeled data. Figure~\ref{fig:wordemb} presents the results. As the unlabeled data becomes smaller, the performance remains unchanged at the beginning and then becomes worse. This demonstrates that the mass of unlabeled data is a key factor to achieve high performance. However, even with only 68MB unlabeled data, we can still observe remarkable improvements (+0.4 on PKU and MSR).
		We also observe that MSR dataset is more robust to the size of unlabeled data than PKU dataset. We conjecture that this is because MSR training set is larger than PKU training set\footnote{There are 2M words in MSR training set but only 1M words in PKU training set.}.
		
		
		We also study how the teacher's performance influence the student. We train other two models that use different teacher models. One of them uses a worse teacher and the other uses a better teacher. The results are shown in Table~\ref{tab:pioneer}. As expected, the worse teacher indeed creates a worse student, but the effect is marginal (-0.1 on PKU and MSR). And the better teacher brings no improvements. 
		These facts demonstrate that the student's performance is relatively insensitive to the teacher's ability as long as the teacher is not too weak.
		
		\begin{table}
			\centering
			\begin{tabular}{l|c|c|c|c}
				\hline
				\multirow{2}{*}{Models}	&	\multicolumn{2}{c|}{teacher}	&	\multicolumn{2}{c}{student}	\\
				\cline{2-5}
				&	PKU	&	MSR	&	PKU	&	MSR	\\
				\hline
				WE-CONV-SEG	&	95.7	&	97.4	&	96.5	&	98.0	\\
				worse teacher	&	95.4	&	97.1	&	96.4	&	97.9	\\
				better teacher	&	96.5	&	98.0	&	96.5	&	98.0	\\
				\hline
			\end{tabular}
			\caption{\label{tab:pioneer} Performances of student models and teacher models. A previous trained model maybe reused in following so that there are some }
		\end{table}
		
		Not only the pretrained word embeddings, we also build a vocabulary $V_{word}$ from the large auto-segmented data. Both of them should have positive impacts on the improvements. To figure out their contributions quantitatively, we train a contrast model, where the pretrained word embeddings are not used but the word features and the vocabulary are persisted, i.e. the word embeddings are randomly initialized. The results are shown in Table~\ref{tab:wordemb}. According to the results, we conclude that the pretrained word embeddings and the vocabulary have roughly equal contributions to the final improvements.
		
		\begin{table}
			\centering
			\begin{tabular}{l|c|c}
				\hline
				Models	&	PKU	&	MSR		\\
				\hline
				WE-CONV-SEG	&	\bf	96.5	&	\bf	98.0	\\
				-word emb	&	96.1	&	97.6	\\
				-word feature	&	95.7	&	97.3	\\
				\hline
			\end{tabular}
			\caption{\label{tab:wordemb} Performances of our models with different word feature options. ``-word emb'' denotes the model in which word features and the vocabulary are used but the pretrained word embeddings are not. ``-word feature'' denotes the model that uses no word feature, i.e. CONV-SEG.}
		\end{table}
		
		\section{Related Work}
		CWS has been studied with considerable efforts in NLP commutinity. 
		\newcite{xue2003chinese} firstly modeled CWS as a character-based sequence labeling problem. They used a sliding-window maximum entropy classifier to tag Chinese characters into one of four position tags, and then coverted these tags into a segmentation using rules. Following their work, \newcite{peng2004chinese} applied CRF to the problem for sequence-level prediction. 
		Recently, under the sequence labeling framework, various neural models have been explored for CWS.
		\newcite{Zheng2013Deep} firstly applied a feed-forward neural network for CWS. \newcite{Pei2014Max} improved upon \newcite{Zheng2013Deep} by explicitly modeling the interactions between local context and previous tag. \newcite{chen2015gated} proposed a gated recursive neural network (GRNN) to model the combinations of context characters. \newcite{Chen2015Long} utilized Long short-term memory (LSTM) to capture long distant dependencies. \newcite{xu2016dependency} combined LSTM and GRNN to efficiently integrate local and long-distance features.
		
		Our proposed model is also a neural sequence labeling model. The difference from above models lies in that CNN is used to encode contextual information.
		CNNs have been successfully applied in many NLP tasks, such as text classification \cite{kalchbrenner2014a,kim2014convolutional,zhang2015character,conneau2016very}, language modeling \cite{kim2015character-aware,Pham2016Convolutional,Dauphin2016Language}, machine translation \cite{meng2015encoding,Kalchbrenner2016Neural,Gehring2016A}. 
		Experimental results show that the convolutional layers are capable to capture more \emph{n}-gram features than previous introduced networks.
		\newcite{Collobert2011Natural} also proposed a CNN based seuqence labeling model. However, our model is significantly different from theirs since theirs adopt max-pooling to encode the whole sentence into a fixed size vector and use position embeddings to demonstrate which word to be tagged while ours does not. Our model is more efficient due to the sharing structure in lower layers. Contemporary to this work, \newcite{Strubell2017Fast} applied dilated CNN to named entity recognition.
		
		The integration with word embeddings is inspired by word-based CWS models \cite{Andrew2006A,Zhang2007Chinese,Sun2009A}. Most recently, \newcite{Zhang2016Transition,Liu2016Exploring,Cai2016Neural} proposed word-based neural models for CWS. Particularly, \newcite{Zhang2016Transition,Liu2016Exploring} utilized word embeddings learned from large auto-segmented data, which leads to significant improvements. Different from their word-based models, we integrate word embeddings with the proposed character-based model.
		
		Simillar to this work, \newcite{Wang2011Improving} and \newcite{zhang2013exploring} also enhanced character-based CWS systems by utilizing auto-segmented data. However, they didn't use word embeddings, but only used statistics features. 
		\newcite{sun2010word} and \newcite{Wang2014Two} combined character-based and word-based CWS model via bagging and dual decomposition respectively and achieved better performance than single model.
		
		\section{Conclusion}
		In this paper, we address the weaknesses of character-based CWS models. We propose a novel neural model for CWS. The model utilizes stacked convolutional layers to derive contextual representations from input sequence, which are then fed to a CRF layer for prediction. The model is capable to capture rich \emph{n}-gram features automatically. 
		Furthermore, we propose an effective approach to integrate the proposed model with word embeddings, which are pretrained on large auto-segmented data.
		Evaluation on two benchmark datasets shows that without any feature engineering, much better performance than previous models (also without feature engineering) is obtained. Armed with word embeddings, our model achieves state-of-the-art performance on both datasets, without using any external labeled data.
		

\section*{Acknowledgements}
This work is supported by the National Key Research \& Development Plan of China (No.2013CB329302). Thanks anonymous reviewers for their valuable suggestions. Thanks Wang Geng, Zhen Yang and Yuanyuan Zhao for their useful discussions.

\bibliography{ijcnlp2017}
\bibliographystyle{ijcnlp2017}

\end{document}